\documentclass{article}
\usepackage[T1]{fontenc}
\usepackage[utf8]{inputenc}
\usepackage{verbatim}
\usepackage{wrapfig}
\usepackage{booktabs}
\usepackage{graphicx}
\usepackage{microtype}

\usepackage{hyperref}
 
\hypersetup{unicode=true, bookmarks=false, breaklinks=false,pdfborder={0 0 0},backref=section,colorlinks=false, hidelinks=true}

\makeatletter

\providecommand{\tabularnewline}{\\}

\usepackage[final, nonatbib]{neurips_2020}

\usepackage[utf8]{inputenc} 
\usepackage[T1]{fontenc}    
\usepackage{hyperref}       
\usepackage{url}            
\usepackage{booktabs}       
\usepackage{amsfonts}       
\usepackage{nicefrac}       
\usepackage{microtype}      
\usepackage{graphicx}       
\usepackage{amsmath}        
\usepackage{multirow}

\title{Relation-weighted link prediction for\\disease gene identification}

\author{
  Srivamshi Pittala\thanks{Equal contribution.}\\
  OccamzRazor\\
  \texttt{srivamshi@occamzrazor.com}
  \And
  William Koehler\\
  OccamzRazor\\
  \texttt{william@occamzrazor.com}
  \And
  Jonathan Deans\\
  OccamzRazor\\
  \texttt{jonathan@occamzrazor.com}
  \And
  Daniel Salinas\\
  OccamzRazor\\
  \texttt{daniel@occamzrazor.com}
  \And
  Martin Bringmann\\
  OccamzRazor\\
  \texttt{martin@occamzrazor.com}
  \And
  Katharina Sophia Volz\\
  OccamzRazor\\
  \texttt{volz@occamzrazor.com}
  \And
  Berk Kapicioglu\footnotemark[1]\\
  OccamzRazor\\
  \texttt{berk@occamzrazor.com}
}
\makeatother

\begin{document}
\maketitle
\begin{abstract}
Identification of disease genes, which are a set of genes associated
with a disease, plays an important role in understanding and curing
diseases. In this paper, we present a biomedical knowledge graph
designed specifically for this problem, propose a novel machine learning
method that identifies disease genes on such graphs by leveraging
recent advances in network biology and graph representation learning,
study the effects of various relation types on prediction performance,
and empirically demonstrate that our algorithms outperform its closest
state-of-the-art competitor in disease gene identification by $24.1\%$.
 We also show that we achieve higher precision than Open Targets,
the leading initiative for target identification, with respect to
predicting drug targets in clinical trials for Parkinson's disease.
\end{abstract}
There are many diseases, such as Parkinson's disease, that have no
known causes or curative therapies, yet afflict millions of people
around the world \cite{DeLau2006}. The preeminent approach to developing
curative therapies for such diseases involves developing drugs that
can alter the disease state by modulating the activity of a molecular
target \cite{Vamathevan2019}. This approach requires successful identification
of such a molecular target, and within the drug discovery pipeline,
this problem is called \emph{disease gene identification}. 

A relatively recent approach for disease gene identification is network-based
drug discovery \cite{Katsila2016,Chen2013}. Network-based drug discovery,
or more generally network biology, assumes that networks can be used
to represent the molecular interactions in human cells and that disease
phenotypes arise due to perturbations of these interactions \cite{Barabasi2011}.
In other words, the phenotypic impact of a mutated gene is not determined
solely by the known function of that gene, but also by how the altered
activity of its gene product spreads throughout the network and affects
the activity of other molecular components that may have no genetic
defects. Within this framework, one can formulate disease gene identification
as a link prediction problem \cite{Liben-Nowell2007,Martinez2016}. 

In this paper, we present a biomedical knowledge graph that represents
various relationships between molecular and phenotypic objects, some
of which are causal relationships between genes and diseases, and
leverage the graph to infer more of such relationships. We propose
a general method to modify link prediction algorithms, which we call
\emph{relation-weighted link prediction}, and demonstrate that our
algorithms outperform existing state-of-the-art (SOTA) in disease
gene identification. We also study the effects of various relation
types on prediction performance. Finally, we show that we achieve
higher precision than Open Targets, the leading initiative for target
identification, with respect to predicting drug targets in clinical
trials for Parkinson's disease. 

\section{Relation-weighted link prediction\label{sec:Relation-weighted-link-prediction}}

Our goal is to identify disease genes, which are a set of genes associated
with a disease. We formulate this task as a link prediction problem
\cite{Liben-Nowell2007,Martinez2016}, where we construct our biomedical
knowledge graph and leverage the information in the graph to predict
previously unknown links between genes and diseases.

A modern approach to solve link prediction is via graph representation
learning (GRL) \cite{Cai2018,Hamilton2017survey}. Historically, given
a machine learning problem on a graph, the main challenge has been
to decide on the most predictive way to encode information about the
graph into a machine learning model. For example, in case of link
prediction, researchers typically encoded the graph using pairwise
attributes between the candidate node pairs, such as the number of
their common friends. However, such hand-engineered features are time-consuming
to design and are not necessarily optimal for the prediction task.
In contrast, GRL studies algorithms that automatically learns how
to encode the graph structure into low-dimensional vectors which are
specifically tuned for the prediction task.

We introduce a simple but important modification to link prediction
algorithms for knowledge graphs called \emph{relation-weighted link
prediction}. Existing link prediction algorithms for knowledge graphs,
such as RotatE \cite{Sun2019}, CompGCN \cite{Vashishth2019}, and
R-GCN \cite{Schlichtkrull2018}, are able to model graphs with more
than one relation type. However, they assign equal importance to all
links regardless of their relation type and do not account for the
imbalances in the number of links across different relation types.
As a result, the model may be biased to predict well on relation types
with more links and poorly on relation types with less links, even
if the relation type of interest has only a small proportion of the
total links. Relation-weighted link prediction is a modification
of existing learning algorithms such that each relation type is assigned
a weight parameter that is optimized to maximize the predictive accuracy
on the relation type of interest, which in our case is \emph{gene-disease}.
In general, relation-weighted link prediction can be used to modify
any heterogeneous link prediction algorithm.

We demonstrate relation-weighted link prediction on RotatE \cite{Sun2019},
a GRL method that has been shown to achieve state-of-the-art performance
on benchmark tasks like link prediction, node classification, and
graph classification. Given an edge of the form $<h,r,t>$, it learns
low-dimensional vectors $\mathbf{h},\mathbf{r},\mathbf{t}\in\mathbb{C}^{k}$
for nodes and relation types such that an edge is assumed to be a
rotation from the head node to the tail node in the complex vector
space.

Equation \ref{eq:rotate_original} shows the loss function of RotatE
for a single edge $<h,r,t>$. It optimizes a distance-based model
with a negative sampling loss 
\begin{equation}
L=-\log\sigma\left(\gamma-d_{r}\left(\mathbf{h},\boldsymbol{t}\right)\right)-\sum_{i=1}^{n}p\left(h_{i}^{'},r,t_{i}^{'}\right)\log\sigma\left(d_{r}\left(\mathbf{h}_{i}^{'},\boldsymbol{t}_{i}^{'}\right)-\gamma\right),\label{eq:rotate_original}
\end{equation}
where, $d_{r}$ is the relation-specific distance function, $\gamma$
is the margin, $\sigma$ is the sigmoid function, $\mathbf{h},\boldsymbol{t}\in\mathbb{C}^{k}$
are the low-dimensional vector representations of $h,t\in V$, and
$<h_{i}^{'},r,t_{i}^{'}>$ is a negative edge which is corrupted from
the original edge.

In equation \ref{eq:rotate_modified}, we modify the original loss
by introducing a relation-specific weight $w_{r}$ that scales the
contribution of the relation type $r$ to the loss
\begin{equation}
L'=-\log\sigma\left(\gamma-w_{r}*d_{r}\left(\mathbf{h},\boldsymbol{t}\right)\right)-\sum_{i=1}^{n}p\left(h_{i}^{'},r,t_{i}^{'}\right)\log\sigma\left(w_{r}*d_{r}\left(\mathbf{h}_{i}^{'},\boldsymbol{t}_{i}^{'}\right)-\gamma\right).\label{eq:rotate_modified}
\end{equation}

\section{Experiments\label{sec:Experiments}}

In this section, we present our experimental results. To prepare for
training and evaluation, we split disease-gene links into train ($80\%$),
validation ($10\%$), and test ($10\%$) sets. During splitting, we
ensure that all nodes in the disease-gene layer occur at least once
in the train set. We then incrementally augment each of the remaining
layers to the train set (e.g. protein-protein, protein-reaction, pathway-disease,
etc.) in order to estimate the contribution of each layer to predictive
accuracy.  More details about the biomedical knowledge graph that
we constructed and its data sources, node types, and relation types
can be found in the Table S1 and S2.

Since our goal is to identify disease genes, when evaluating disease-gene
links in the test set, we treat all possible genes as candidates for
the disease. We use two well-known link prediction metrics, \emph{filtered
hit@k} and \emph{filtered mean rank (MR)} \cite{Shi2017}, to measure
predictive accuracy. We also introduce a new metric called \emph{filtered
mean percentile (MP)}, which is mean rank normalized by the number
of candidates, to enable a fair comparison between test sets where
the number of gene candidates differ.

For hyperparameter optimization, we use a library called Optuna \cite{Akiba2019}.
Optuna allows us to sample the hyperparameter space efficiently and
prune unpromising runs early. We optimize on the validation set and
report results of the best performing model on the test set.

\subsection{Relation layers}

Here, we demonstrate how augmenting the graph with new relation layers
and modifying the objective function affects prediction performance.
For these experiments, we let valid and test splits consist only of
disease-gene edges, which we keep constant across experiments. The
only thing we change between experiments is the new relation layers
we augment to the train split. As the link prediction algorithm, we
use the relation-weighted modification of RotatE.

Results are shown in Table \ref{tab:doidgenet_layers}. Due to lack
of space, we abbreviate curated DoidGeNET as DG, uncurated DoidGeNET
as DG\_UC, Disease Ontology as DO, and Reactome as RT. We observe
that each relation layer we add to the train split improves the prediction
performance compared to the preceding one, and the fully augmented
graph achieves the best performance. This shows the benefits of adding
relevant biological information to the graph. Furthermore, in Table
\ref{tab:doidgenet_unmodified_vs_modified}, we compare the prediction
performance of the original and the relation-weighted RotatE on the
fully augmented graph. We observe that the relation-weighted variant
performs better than the original, thus showing the utility of weighting
relation types in heterogeneous graphs. More details about the optimal
relation weights learned by the model can be found in the Table S3.

In conclusion, when we augment our graph with all of the relation
layers and apply our relation-weighted objective, relative to the
graph that only consists of the disease-gene layer, we achieve a relative
reduction of $76.2\%$ in MR, a relative increase of $98.4$\% in
hit@30, and a relative increase of $28.2\%$ in MP. 

\begin{table}
\centering{}\caption{\label{tab:doidgenet_layers}Contribution of relation layers to prediction
performance}
\begin{tabular}{|l|c|c|c|}
\hline 
Variant & hit@30 & Mean Rank & Mean Percentile\tabularnewline
\hline 
\hline 
DG & 0.189 & 4995.65 & 72.77\tabularnewline
\hline 
DG + STRING & 0.287 & 2029.74 & 88.94\tabularnewline
\hline 
DG + STRING + DG\_UC & 0.353 & 1467.84 & 91.64\tabularnewline
\hline 
DG + STRING + DG\_UC + DO & 0.363 & 1256.69 & 92.84\tabularnewline
\hline 
DG + STRING + DG\_UC + DO + RT & \textbf{0.375} & \textbf{1186.81} & \textbf{93.32}\tabularnewline
\hline 
\end{tabular}
\end{table}

\begin{table}
\centering{}\caption{\label{tab:doidgenet_unmodified_vs_modified}Comparison of original
and relation-weighted RotatE on the full graph}
\begin{tabular}{|l|c|c|c|}
\hline 
Variant & hit@30 & Mean Rank & Mean Percentile\tabularnewline
\hline 
\hline 
Original & 0.368 & 1298.44 & 92.70\tabularnewline
\hline 
Relation-weighted & \textbf{0.375} & \textbf{1186.81} & \textbf{93.32}\tabularnewline
\hline 
\end{tabular}
\end{table}

\subsection{Comparison with state-of-the-art (SOTA)}

We compare our best performing model against existing SOTA methods
for disease gene prediction \cite{Agrawal2018}. Specifically, we
compare against direct neighborhood scoring \cite{Navlakha2010},
DIAMOnD \cite{Ghiassian2015}, and random walks \cite{Zhou2016,Leiserson2015}. 

Direct neighborhood scoring \cite{Navlakha2010} assigns each gene
a score that is proportional to the percentage of its neighbors associated
with the disease. To construct disease gene clusters, it initializes
the clusters with a seed set of disease genes and recursively expands
the cluster with the highest scoring genes. DIAMOnD \cite{Ghiassian2015}
also initializes clusters with seed genes, but uses a statistic called
\textit{connectivity significance} to expand them. Random walks \cite{Zhou2016,Leiserson2015}
use seed genes to initialize a random walker that randomly visits
neighbor genes. Upon convergence, the frequency with which the nodes
in the network are visited is used to rank the disease genes.

For a fair comparison, we initialize all these methods with the same
seed genes that were used to train our own models. When appropriate,
we optimize their hyperparameters using the validation set. Because
of the way these methods are designed, we train them on a subgraph
consisting of two layers: disease-gene (DoidGeNET) and protein-protein
interaction (STRING).

In Table \ref{tab:doidgenet_sota}, we share the results we obtained
on the test set\footnote{Some cells are NA because they were too expensive to compute.}.
In conclusion, relation-weighted RotatE outperforms all SOTA methods
for disease gene prediction, including its closest competitor DIAMOnD.
Specifically, compared to DIAMOnD, our model achieves a relative increase
in hit@30 of $11.6\%$ and a relative increase in hit@100 of $24.1\%$.
\begin{table}
\begin{centering}
\caption{\label{tab:doidgenet_sota}Comparison with SOTA methods}
\begin{tabular}{|c|c|c|c|c|}
\hline 
Method & hit@30 & hit@100 & Mean Rank & Mean Percentile\tabularnewline
\hline 
\hline 
Relation-weighted RotatE & \textbf{0.375} & \textbf{0.535} & \textbf{1186.81} & \textbf{93.32}\tabularnewline
\hline 
DIAMOnD & 0.336 & 0.431 & NA & NA\tabularnewline
\hline 
Direct neighborhood scoring & 0.250 & 0.357 & 3339.61 & 80.24\tabularnewline
\hline 
Random walk & 0.007 & 0.026 & 4597.91 & 72.78\tabularnewline
\hline 
\end{tabular}
\par\end{centering}
\end{table}

\subsection{Comparison with Open Targets}

\begin{figure}
\centering{}\includegraphics[scale=0.5]{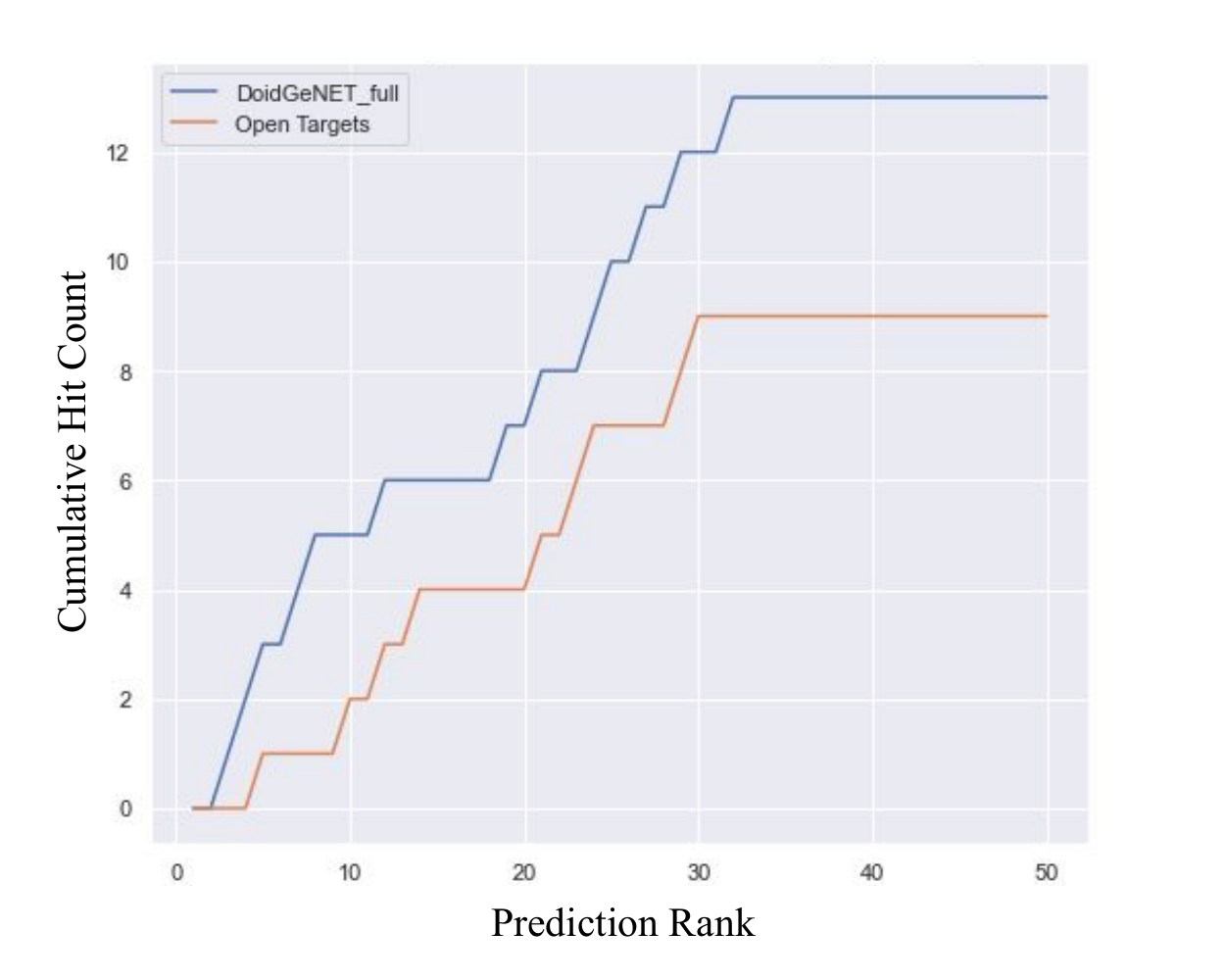} \includegraphics[scale=0.5]{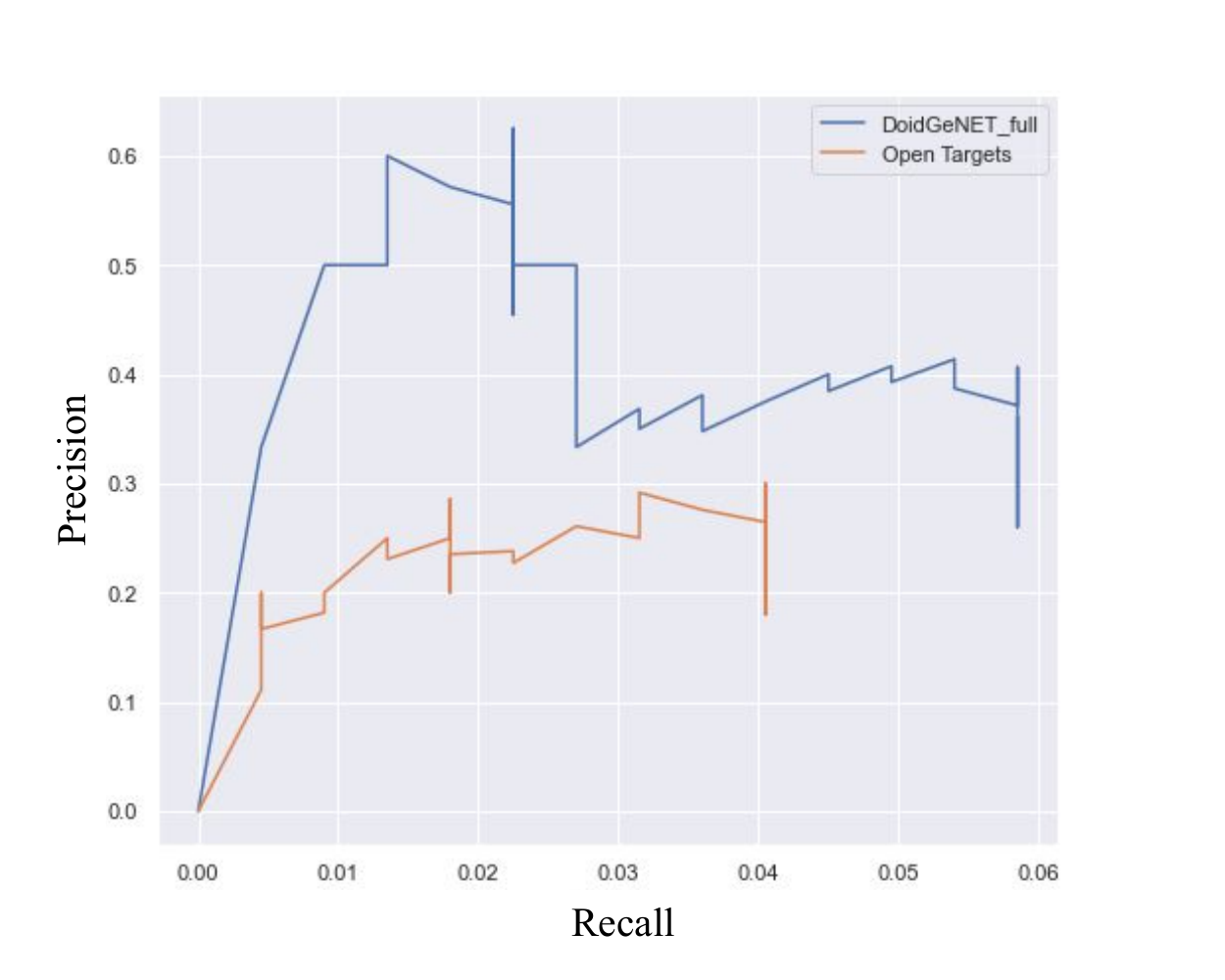}\caption{\label{fig:open_targets_pd}Comparison between our model and Open
Targets for predicting drug targets in clinical trials for Parkinson's
disease. Left: cumulative hits in the top 50 predictions. Right: precision
versus recall plot for top 50 predictions.}
\end{figure}
To further assess the practical utility of our approach in identifying
disease genes, we compared our predictions against those of Open Targets
\cite{Carvalho-Silva2019}, the leading initiative for target identification,
with respect to predicting drug targets in clinical trials for Parkinson's
disease. To obtain the list of targets in clinical trials, we used
Trialtrove \cite{Zarin2011}, the most comprehensive database for
clinical trials. To ensure a worst-case comparison, we removed all
genes that appeared in the evaluation data from the training data
of our models, but we were not able to do the same for the Open Targets
model, which obtained an unfair advantage because of that.

In Figure \ref{fig:open_targets_pd}, we show a comparison between
the top 50 prediction from our best performing model (i.e. relation-weighted
RotatE trained on the fully augmented graph) and Open Targets. We
observe that, in the top 50 predictions, our model finds more candidates
than those listed by Open Targets, while also consistently achieving
higher precision than Open Targets. 

\section{Conclusion\label{sec:Conclusion}}

In this paper, we presented a biomedical knowledge graph designed
specifically for disease gene identification, proposed a novel machine
learning method that identifies such genes by leveraging recent advances
in network biology and graph representation learning, studied the
effects of various relation types on prediction performance, and empirically
demonstrated that our algorithms outperform its closest state-of-the-art
competitor in disease gene identification by 24.1\% . We also showed
that we achieve higher precision than Open Targets, the leading initiative
for target identification, with respect to predicting drug targets
in clinical trials for Parkinson's disease.


\bibliographystyle{unsrt}
\bibliography{gene_disease, gene_disease_manual}

\end{document}